\pdfoutput=1
\documentclass[11pt]{article}
\usepackage[nohyperref]{acl}
\usepackage{times}
\usepackage{graphicx}

\title{ Communication, Vocalisation, and Language }
\author{Peter Wallis \\
  pwallis@acm.org}

\begin{document}
\maketitle

\begin{abstract}
The contention is that the current batch of Large Language Models don't \textit{understand} language, but they \textit{do} it.  Rather than LLMs forming some kind of ``distributed representation'' of symbols, the claim is that LLMs are capturing ``Cultural Intelligence'' and grounding symbols in what the sociologists call ``practices''.
\end{abstract}

Human communication classically involves vocalisation and is definitively language-like, but it can of course happen without vocalisation as writing, and we can speak without communicating - talking to one's self for example.  We can also communicate without vocalising or language. Consider two people moving a sofa.  How and where to move the sofa might be initiated with speech, but the details are communicated by \textit{interaction}. Both lift the sofa, one starts moving, and the other moves in the same direction. One stops, and the other pauses. Communication happens, but without vocalisation and without language.  Figure 1 shows these three aspects as dimensions to human communication.

The notion of language is of course contested, but here I want to consider a ``pure'' language in the Fregean sense in which symbols are used to refer to stuff in the world.   Computers are really good at manipulating signs, and can perform arbitrary transformation on the surface form of token sequences, but how exactly do signs get meaning? How are the signifieds chosen and specifically, when a signified is a set of things (a type) what determines set membership?  In English the symbol 'cat' has a \textit{good enough} meaning for us to refer to cats in context, but how does that work?  The signified is the set of all cats as English speaker's know them, but does that set of animals include the big cats? How about dead cats? If symbol manipulation is an accurate model of the mind, then how, exactly, is the mind calculating ``good enough''?  In classical AI, pragmatics considers how meaning in context is reified but it is usually done in terms of using other symbolic representations. This paper is within the enactivist movement~\cite{varieties17} and follows up on Wittgenstein's notion of meaning as use~\cite{WgsnBB}.
\begin{figure}
\caption{\textit{A Venn diagram of language, vocalisation, and communication with human communication in the middle.} }
\includegraphics[height=60mm]{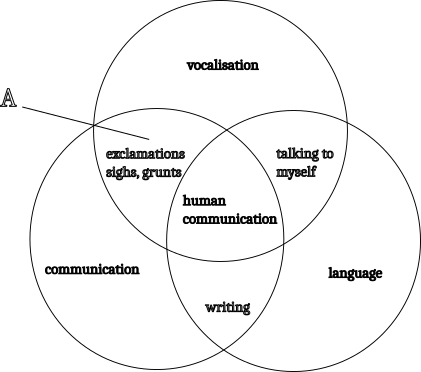}
\end{figure}

The early twentieth century was an turbulent time and while Wittgenstein was thinking about the limitations of formal logic, Vygotsky and colleagues were inventing what would become Soviet psychology.  A key observation was that there is ``a pre-linguistic stage in the development of thought'' and a ``a pre-intellectual stage in the development of speech''.  Over time these two merge to form adult thinking~\cite{kozulin90}.  In the next section I propose an enactivist interpretation of Wittgenstein's discussion of language games which aligns with the notion of ``a pre-intellectual stage in the development of speech''.  If meaning is based on use however, vocalisation needs to be used \textit{before} words have meaning. That is, we need to look at communication in region A of Figure 1.

Section 3 introduces the notion of \textit{practices}~\cite{practices}.  The idea will look familiar to computer scientists familiar with Schank and Abelson's scripts~\cite{schank89}, but rather than thinking of practices as involving \textit{things}, practices provide a model of ``primitive thinking'' -- a pre-linguistic form of thought. The proposal is that Large Language Models are doing this rather than forming distributed representations. The claim is that conscious thought is indeed symbolic and based on natural language, but there is another type of speaking and thinking.

\section{Pre-intellectual Speech}

In the Brown Book, Wittgenstein introduces a made up language that is used by a community of stone masons. The entire conversation is about bricks.  The child learns the language from grown-ups by being trained to its use.  He says the word ``trained'' here is being used in a way that is strictly analogous to that in which we talk of an animal being trained to do certain things.  Rather than reward and punishment being used to have a dog respond appropriately to the word ``fetch,'' language learners in this community are taught to bring a block when the brick layer calls out ``block!'', and a slab when the brick layer calls ``slab!''.  Wittgenstein characterises this as a technical language that is, however, complete. To quote:
\begin{quote}
 Systems of communication [like this] we shall call ``language games''. They are more or less akin to what in ordinary language we call games. Children are taught their native language by means of such games, and here they even have the entertaining character of games.  We are not, however, regarding the language-games which we describe as incomplete parts of a language, but as languages complete in themselves, as complete systems of human communication.
\end{quote}

So how might the child be playing this game?  Wittgenstein suggests that the sounds are enough.  Rather than the sign ``slab'' \textit{referring to} a brick in the world with a particular shape, to the child there is just the sound - a sound that can be differentiated from the sound ``block''.  With a bit of reinforcement learning and a teacher, the sound can be associated with a (previously established) behaviour involving carrying bricks.  There is vocalising, and there is communication, but the sign ``slab!'' doesn't need to signify a slab for the child. 

Of course from our perspective -- and presumably that of the brick-layer -- there are indeed things called ``slab'' and ``block'' and these signs refer to slabs and blocks whether the child knows the meaning or not, but what is the child doing?

\subsection{Wittgenstein's child}

The brick-layer as an enculturated adult \textit{knows} there is a set of bricks that are called ``slabs'', but the child, in the early stages, doesn't know that.  The child can however \textit{detect} slabs through the mechanism for interacting with bricks.

In philosophy they point out that the term ``Evening Star'' does not \textit{mean} Venus, even if that is what the term references.  Before discussions of meaning however, note how an Evening Star \textit{detector} could conceivably be made using a telescope, an LDR (Light Dependent Resistor) and a bit of circuitry.  When the sun goes down and the first star of the evening becomes visible, this device might turn on an indicator light.  Does the indicator light \textit{mean} ``Venus is there''?  Well it does to a human observer who has faith in the existence of planets, but the most that could be said for the device itself is that the indicator light means light is falling on the LDR.  Any referential meaning is strictly internal to the device itself.  The conclusion is that the referential meaning of a sign is to a model of the world, and the model faces the world as a totality.  The term ``Morning Star'' assumes a model of the world in which there are things called stars that appear at night and one is the last to disappear in the morning and another, the Evening Star, is the first to appear in the evening.  The term ``Venus'' assumes a model in which planets circle the sun, and one is close enough to the Earth to produce the same phenomena.

The observation is that, at some point, the child doesn't have a model.  For the child, at some point, there is just the phenomena.  What the child does have however is a brick detector that enables the activity of carrying bricks.  There is an affordance in the shape of the brick that ``is a fit'' with the child's hands.  The child can pick up and carry slabs and other bricks before the formation of categories of brick.  The child's SDR - Slab Dependent Resistor - is of course not purely visually nor indeed strictly sensor based.  The child, before having language of any kind, external or internal, interacts with the world and learns how to recognise which bits are probably pickable-up-able and which are not.  That is, before the world is thought of as having things in it, the child has detectors that indicate the possibility of further action. Successful further action confirms the detector as useful.  To \textit{recognise} something is to map from a set of phenomena to an entity in a model.  A \textit{detector} reacts to phenomena.  Of course a detector might act as the front end to a recogniser, but 70 years of AI research suggests that nearly always the model is required to provide top down information about possible explanations for partial sensor data.

In summary, Wittgenstein's child starts with detecting phenomena and participates in games that teach the child a model of the world in which there is not just phenomena, but \textit{things}. Things that can be referred to with words.  This will eventually enable the child to think of the world in terms of blocks and slabs, wedges and pyramids, and to imagine constructions such as towers that are made of these components.  Before that however, the child can successfully interact with the world and others using just the phenomena. But there's more.  It turns out that an agent can often reason without symbolic representation.

\section{ Pre-linguistic Thought}

As adults we think about things.  We look at a pile of bricks and think ``I could build a tower using three blocks and a pyramid''.  In Vygotsky's terminology we think ``scientifically'' but if the child is not thinking about blocks and slabs, how do they do planning?  How would ``a pre-linguistic stage in the development of thought'' work?

Kozulin, in his description of the early days of Soviet psychology, provides this vignette care of Lula:
\begin{quote}
An illiterate peasant from a remote village was shown a hammer, a saw, a log, and a hatchet and asked to tell which of these objects could be called by the same word.  At first the peasant rejected the very idea of a common name, arguing that you cannot call different objects by the same name. When prompted by the suggestion that ``one fellow said the hammer, saw and hatchet are alike'', the subject responded that the log also belongs in the same group.  When the experimenter suggested the hammer, saw and hatchet are tools, the subject responded: ``Yes, but even if we have tools, we still need wood -- otherwise, we can't build anything.
\end{quote}

The conclusion was that children (and those without the benefit of being educated in the ways of scientific thinking) could get by very well with reasoning in terms of \textit{practices}.  The peasant has the sensory experience of seeing a hammer, and what is called up is the practice of building something.  One might object that the practice of building something requires not just the recognition, but also  (symbolic) representation of hammers, but as described above, there is plausibly just correlation between sensory data and the activity, reinforced with training.  Shown a hammer, what our peasant conjures up is not a set of things which are hammer-like, but the activity of doing something with a hammer.  Hearing the word ``hammer'', he doesn't think of hammers, but of the practice of building stuff.

Language is not good at capturing what is going on here. The contention is that language ``thingifies'' reality and hence is quite poor at describing actions without introducing the things involved.  Consider walking down the street and seeing a man at a bus stop.  As described, you know what the man is most likely doing; he's catching a bus.  But to convey the situation and the phenomena available to your senses, I have had to use words that reference things -- ``street'', ``man'', and ``bus stop''. Not only do we need words to describe the activity, we need words to think (consciously) about it.  Hutto (see \cite{gallagher2020}) argues that when we don't think about it, we still \textit{directly perceive} intentions.  Walking down the street, what we see is a bus-catching activity performed by a man. Like Vygotsky's peasant and the hammer, our sensory experience directly conjures up the practice of catching a bus.  The idea is that human nature is based in a field of practices where each \textit{practice is an array of activities} that is public and identifiable~\cite[p.~11]{practices}.

I interpret the term ``activity'' here as others have used the term ``behaviour''.  Brooks' robots \cite{brooks91} -- the Roomba being the classic -- have a set of behaviours that are reflexive and produce situated action by connecting actuators directly to sensors.  The challenge is to choose which behaviour to do.  Arkin's ``Behaviour-based Robotics''~\cite{Ark98} uses classic AI reasoning with symbols and a model of the world to choose which to activate or inhibit in order to achieve goals -- goals represented with symbols.  Where as ``behaviour'' suggests an observer is present, the notion of ``activity'' emphasises unconscious doing.  The Roomba does the activity of spot cleaning or charging the battery, and the child picks up bricks or carries bricks. These activities do not require a theory -- there is no need for a model of the world -- ``the world is it's own best representation'' to quote Brooks. What an activity does need is an ``embodied'' agent~\cite{HeJa96,wooldridge23} that can do things in an environment, but that is all.

The observation is that having the embodied agent remember a \textit{sequence} of activities is a game changer.  Wittgenstein's child starts by learning the activities for carrying blocks and slabs and so on, but the interaction with the brick layer is not random. The brick layer's requests have a structure and if the child can remember the sequence of activities performed, then that structure becomes susceptible to statistical analysis.  If we assign a token to each performance of an activity, and then store the sequence of tokens, a markov model can predict which activity will occur next.  Such tokens, like the indicator light in the Venus detector example above, have ``meaning'' only in that they reference something about the agent itself. A markov model or artificial neural network can predict which activity the embodied agent will be performing next.  The brick layer might set out with the goal of building a tower made out of things called blocks and pyramids, but with enough towers built, the robot child may simply remember the sequence of the carrying-a-block activity (times 3) plus (1 off) carrying-a-pyramid.  If the brick-layer has asked for three blocks, the child might reasonably pick up a pyramid.  The brick-layer, being a symbol using adult, might suspect the child knows his intention and might assume the child shares his goal of building a tower.  The child's brain however might be just doing the stats.
For those familiar with the history of AI, this no doubt looks like case-based reasoning (see \cite{cbr}, \cite{rpd} and \cite{mpf}). The key advance here is to run the mechanism over an embodied agent' inherent activities rather than action descriptions produced in an armchair.

Vygotsky's interest was in child development in the presence of adults, in a ``zone of proximal development'' or ZPD.  The take away message here is that the child becomes quite good at predicting what the brick-layer will want next, without thinking about bricks or other minds. The child can participate in communication \textit{before} there is any notion of things with labels or any notion of the brick-layer's beliefs and intentions. The brick-layer, as an enculturated adult, makes the mistake of thinking the child thinks like us and, in the process, teaches the child to think like us.

We make the same mistake with animals.  No representation of a future state of the world is required in order for a dog, an ape, or a child to ``read the mind'' of a carer when they pick up the lead, open the cage, or pick up the car keys.  Vervet monkeys screech differently when they see a snake to when they see a cheetah, and this is seen as evidence that they are representing cheetahs and snakes.  An alternative would be that the monkeys have an established behaviour for dealing with snakes, and another for dealing with cheetahs.  Rather than individuals being trained to associate vocalisations with behaviours, natural selection can make the link.  To us humans one screech might mean cheetah, but for the monkey, the screech ``conjures up'' the climb-a-tree behaviour.  There is plausibly no abstraction and no metaphysics, just correlation between the vocalisation and practice.

The notion of object permanence is equally problematic. A robot child with an ADR - ambulance detecting resistor - and a bit of memory can learn to predict that,  if its ADR goes off pointing out one window, moments later it will go off if it turns to look out the other window. The system does not need a model of the world in order to predict the behaviour of ambulances, and Wittgenstein's child does not need a model of the brick-layer's mind nor to have ``an image-like representation of the future state of the world'' (see \cite{tomasello08}).  There is no need for a ``cognitive model'' and no need for a ``language of thought''.

When it comes to contemporary AI and the Large Language Models, the proposal is that,rather than forming (distributed) symbolic representations of things and reasoning ``scientifically'',  LLMs are implementing ``primitive'' thinking and using ``glorified auto-complete'' (see Hinton in ~\cite{GAC}) over representations of practices.  Because practices are grounded in activities, and activities are internal to the system, these systems do not have a symbol grounding problem.

\section{The Big Question}

Words introduce the idea that the world is not just a soup of phenomena; using words to communicate introduces the \textit{theory} that the world is composed of things.  This is a really useful. Thinking of the world as composed of (atomic) things, we make models to explain the observable phenomena.  We introduce ``gravity'' to explain the behaviour of the (mobile) stars, and we introduce the ``subconscious'' to explain irrational behaviour. We also introduce beliefs and desires to explain other peoples (rational) behaviour.  The way we carve the world~\cite{wfdt} is not however arbitrary~\cite{culSau}.  Although the number of categories for types of snow might indeed be theoretically infinite and the choice is indeed a cultural product reflected in the culture's language, the carving is grounded in human activity.  Diverse cultures divide the world in roughly similar ways because human activities are broadly universal.

So, vocalisation introduces discernible words, and the teaching of words teaches the infant to consider the world as a collection of things with attributes and relations between them.  This is a really useful tool.
Vygotsky and his colleagues went on to look at how children move from ``primitive'' thinking to reasoning as we know it.  Cultural Psychology considers intelligence as we think of it as being a skill maintained and developed by a community - a community of practice.  Glorified Auto Complete might be (far) more useful than it sounds, but thinking in terms of a world composed of things is even more useful.
 Over time the next generation are taught more ``cognitive tools'' that have proved effective.  As children, we are taught the sequence ``one-two-three'' and on to twelve, and we are taught how to align this ``song'' with the activity of putting apples in a box, or pointing at kittens in a basket. We are taught the times tables, and how to use them.  We are taught to respect our elders and about the roles individuals play in society.  We are taught that there is indeed something called a ``bus stop'' and you can use it not only to perform the activity of catching a bus, but also to put up posters, or as a place to meet up.

\subsection{LLMs and the future}

In a previous paper~\cite{wallis24} I proposed that LLMs are not reasoning scientifically, but rather performing the (socially situated) activity of \textit{convincing}. Of course what LLMs can and can't do is a moving target but here I hope to provide a framework for thinking about how we could, and if we should, make them work better.  The first proposal is that LLMs are indeed capturing ``cultural intelligence'' from all the data they are trained on. When things go wrong, it might be because they do not get the embedded dimension to being human.  Basically, they do not pick up slabs and blocks, and they think that glue on pizza is a good idea because, well, they have never tasted glue.  Quoting Rik Mayall talking about ear wax, ``..and we all know what that tastes like don't we boys and girls.''  It might be forty years since you tasted ear wax, but the thought still conjures up the experience.  Similarly for glue.  There are no doubt edible glues that don't taste like glue, but that is not our experience.  The basis of common-sense does not need to be rational, and it does not need to be recent.

A second failure mode would be when a cognitive tool is not available, or they don't know the sequence-of-activities to use it.  At one point the LLMs were not good at big number mathematics.  They could say what two plus three is, but not what twenty one million, four hundred and twelve thousand, seven hundred and seventeen plus three is. The practice of using (rote learnt) small number maths to sum big numbers was not there. Similarly an LLM might know that Charles is a king, and know how to help customers access a corporate website, but have no idea how to help King Charles use an NHS website without pressing all kinds of conversation analytic buttons~\cite{AlbertStokoe26}.  Rather than AI being good at everything, but needing some ``guard rails'' to constrain it (suggesting of course that the guardrails are someone else's problem -- the government's presumably) failures might be caused by missing cognitive tools.  Where do we find missing tools? By studying what people do and how they solve problems in the cultural context. I propose the tools for doing this are the ethnographic variants of Conversation Analysis~\cite{tHave99}.

Vygotsky died at 37.  He didn't have much time and knew it. Between that and the fact that he was a clear thinker and prodigious writer, he certainly wanted to attack the big issues and the writing that survives certainly indicates his interest in the nature of consciousness.  In retrospect we can perhaps get some idea of where his thinking was headed.  Thingification teaches us that experiential phenomena don't just come in recognisable sequences, there are things and they have object permanence. We are also taught that the self is an object with permanence.  Glorified auto complete works as a reasoning mechanism when there is a memory of the consequences of a sequence, but for the worm, the cat, the monkey and indeed the toddler, the concept of death is not something available for reflection because there is no prior example available showing the consequences.   Perhaps a notion of the self as an object with permanence is not something we want to give these systems, but then perhaps it is too late.

\section{Conclusion}

Speech as humans use it can -- and often does -- perform communication without needing symbols grounded in reference to things.  What is more Vygotsky's 'primitive thinking,'  conceived as the 'completion model' of LLMs, doesn't need a similarly grounded 'language of thought'.  Yes, the LLM mechanism works over tokens, but those tokens are conceivably grounded in relationship between an agent and its environment with tokens representing activities, and sequences of tokens representing (successful) practices.  An agent can predict the future, not by modelling the world, but by simply remembering what happened last time. 

Thinking of the world as composed of things and their relationships is however an incredibly powerful cognitive tool, passed on to the next generation through learning to use language.  Rather than words doing things in a social context, words become symbols referring to things in a world model. Grounding becomes the process of connecting the signs with the signified and in the process ``thingifies'' our phenomenological experience of the world. Phenomena become evidence for the existence of a thing, and we think ``scientifically'' in terms of the world model.
The suggestion is that we should look closely at how we raise children to be good members of society and look at how that process should, or should not, be applied to LLMs.

\bibliography{../../mybib.bib}

\end{document}